\documentclass[preprint,12pt,authoryear]{elsarticle}

\usepackage{tabularx}
\usepackage{amssymb}
\usepackage{amsmath}
\usepackage{graphicx}
\usepackage{booktabs}
\usepackage{hyperref}
\usepackage{multirow}
\usepackage{rotating}

\begin{document}

\begin{frontmatter}

\title{Broadening Access to Transportation Safety Data with Generative AI: A Schema-Grounded Framework for Spatial Natural Language Queries}

\author[umass]{Mahdi Azhdari\corref{cor1}}
\ead{mazhdari@umass.edu}

\author[umass]{Eric J. Gonzales}

\affiliation[umass]{organization={Department of Civil and Environmental Engineering, University of Massachusetts Amherst},
            city={Amherst},
            state={Massachusetts},
            country={USA}}

\cortext[cor1]{Corresponding author.}

\begin{abstract}
Transportation safety analysis requires integrating crash records, roadway attributes, and geospatial data through GIS-based workflows, but access remains uneven across agencies and community stakeholders. Technical prerequisites create a gap between analytical tools central to safety planning and the practitioners able to use them. Local agencies, school committees, and residents may have safety concerns but limited capacity to retrieve, filter, map, and analyze relevant data. Generative AI offers a way to narrow this divide, but its public-sector use raises questions about reliability, reproducibility, and governance. This paper presents a schema-grounded natural language interface for transportation safety analysis, using a large language model (LLM) to interpret user intent while preserving deterministic, reviewable execution against an authoritative database. User queries are translated into structured semantic frames, validated by a rule-based layer, compiled into a typed directed acyclic graph of spatial operations, and executed against a PostGIS database. This bounded design separates language interpretation from deterministic execution, keeping results reproducible and schema-grounded while removing access barriers. The framework is evaluated using a statewide Massachusetts transportation safety database integrating crash records, roadway attributes, and geospatial layers including schools, bus stops, crosswalks, and municipal boundaries. All queries executed successfully; the validation layer corrects errors in 29\% of evaluation queries, reflecting the gap between flexible natural language and strict schema-grounded requirements. The results suggest that combining natural language accessibility with deterministic execution is a practical direction for broadening access to transportation safety data, with implications for trustworthy AI in public-sector planning.
\end{abstract}


\end{frontmatter}
\section{Introduction}
\label{sec:introduction}

Transportation safety analysis increasingly relies on combining crash records, roadway and infrastructure data, and spatial methods to support screening, prioritization, and policy decisions. Agencies use these analyses to identify high-risk corridors, assess conditions near schools and transit stops, compare jurisdictions, and guide the allocation of limited safety resources. In practice, however, conducting this work requires technical familiarity with geographic information system (GIS) platforms, database querying, and the structure of underlying safety datasets, prerequisites that create a gap between the analytical tools now central to transportation safety planning and the range of practitioners able to use them directly. This gap affects municipalities, planners, school safety committees, and community advocates and members: each may have clear safety concerns and legitimate needs for structured transportation safety evidence, whether for infrastructure requests, funding applications, or local advocacy, yet lack the technical knowledge to retrieve, filter, join, aggregate, and map the relevant data. When obtaining this evidence depends on specialized workflows, even straightforward safety questions can be costly to answer, resulting in delays or remaining unanswered. The challenge is therefore not only technical but also institutional, because the ability to conduct structured safety analysis shapes who can participate in safety planning and whose concerns are translated into actionable evidence.

Recent advances in large language models (LLMs) offer a potential way to narrow this divide. Natural language (NL) interfaces can make structured data systems more accessible by allowing users to express analytical intent directly without requiring familiarity with GIS platforms or query languages. But making safety data queryable is only part of the problem; the results also need to be reproducible and trustworthy enough to support real planning decisions. Most existing LLM-based geospatial work has focused on general-purpose queries, agentic execution, or code generation, with relatively little attention to the institutional requirements related to transportation safety planning. In this context, systems must support flexible queries that are reproducible, consistent, and aligned with established analytical workflows.

This paper contributes an NL interface that uses an LLM as a controlled interpretation layer within a structured transportation safety analysis framework. User queries are translated into structured semantic frames, validated and corrected against a domain-specific schema, and compiled into a typed directed acyclic graph (DAG) of spatial operations executed against an authoritative spatial database. This design allows users to express analytical intent in plain language while maintaining schema-grounded, reproducible, and auditable execution. The goal is not to replace established safety analysis workflows but to make them more accessible across a broader range of institutional and community users, including those without technical GIS expertise, while keeping execution bounded and subject to institutional oversight.

The system is developed and evaluated using a statewide Massachusetts transportation safety database integrating crash records, roadway attributes, and geospatial layers such as schools, bus stops, crosswalks, and municipal boundaries. It supports structured safety analysis across multiple contexts while producing outputs such as interactive maps, ranked tables, and exportable datasets. The paper also discusses how this approach can help narrow the gap between the technical demands of transportation safety analysis and the broader range of stakeholders who can benefit from these analyses.

The remainder of the paper is organized as follows. Section~\ref{sec:background} reviews related work on data-driven safety practice and GIS access barriers, natural language interfaces and LLM-based query systems, and trustworthiness considerations for AI in public-sector planning. Section~\ref{sec:architecture} presents the system architecture. Section~\ref{sec:evaluation} presents the evaluation design and results. Section~\ref{sec:discussion} discusses applications, trustworthiness considerations, and future directions. Section~\ref{sec:conclusion} concludes the paper.

\section{Background and Related Work}
\label{sec:background}

\subsection{Transportation Safety Analysis and GIS Access}
\label{subsec:bg_safety}

Transportation safety analysis in the United States is increasingly shaped by data-driven frameworks established through federal safety programs. The Highway Safety Improvement Program (HSIP) requires agencies to systematically identify crash problems, prioritize locations for intervention, and evaluate safety outcomes \citep{fhwa_hsip_manual}. Complementing this, systemic safety approaches extend beyond historically high-crash locations to identify roadway characteristics associated with elevated risk across broader networks \citep{khan2024advancing, fhwa_systemic}. Together, these frameworks rely heavily on the integration of crash records, roadway attributes, and geospatial infrastructure data through GIS-based analysis. Spatial methods such as hotspot detection, proximity analysis, and infrastructure-linked screening have become common tools for identifying safety concerns around schools, transit stops, corridors, and other transportation environments \citep{oke2025bus, fhwa_gis_statedots, mohammed2023gis}.

Despite the growing sophistication of these analytical methods, access to them remains uneven. Prior assessments of GIS use in transportation safety have identified persistent barriers related to technical expertise, data integration complexity, and organizational capacity, particularly for smaller agencies and local stakeholders \citep{fhwa_gis_needs, GUO2020100091}. These barriers extend beyond formal institutions: community groups, neighborhood advocates, and residents seeking to document safety concerns or support requests for infrastructure investment face the same analytical challenges, often without the organizational resources to address them \citep{mcdonald2013assessing}. While many planning and policy questions are conceptually straightforward, translating them into structured analytical workflows often requires familiarity with GIS platforms, database systems, and local data schemas. As transportation agencies increasingly move toward data-driven planning, improving access to these analytical capabilities remains an important practical challenge.

\subsection{Generative AI and Natural Language Access to Transportation Data}
\label{subsec:bg_nlp}
Recent advances in LLMs have created new opportunities to reduce these barriers. Across transportation, generative AI applications have largely focused on traffic operations, autonomous systems, prediction, and simulation \citep{da2025generative, maksoud2025applications, nie2025exploring}. More recently, attention has begun shifting toward the use of LLMs as interfaces for structured analytical tasks. This broader movement aligns with research in natural language interfaces to databases (NLIDBs), which seek to translate user questions into structured database queries. Building on earlier rule-based systems \citep{androutsopoulos1995natural}, modern text-to-SQL approaches increasingly leverage LLMs to improve schema-aware query generation \citep{gao2023text}, while extensions to spatial and spatio-temporal databases further broaden this paradigm \citep{redd2025queries}. Transportation safety analysis, however, involves domain-specific entities, field structures, and geographic conventions that general-purpose query systems are not typically designed to handle consistently, including proximity-based screening near locations such as schools or transit stops, infrastructure-linked filtering, and program-specific temporal analysis.

\subsection{Geospatial AI Systems and Trustworthiness in Public-Sector Contexts}
\label{subsec:bg_gis}

Parallel developments in LLM-enabled GIS systems have further expanded the role of generative models in spatial analysis. Systems such as Autonomous GIS \citep{li2023autonomous}, LLMFind for geospatial data retrieval \citep{ning2025autonomous}, GIS Copilot for spatial analysis \citep{akinboyewa2025gis}, and related geospatial agents increasingly use natural language interfaces to broaden access to spatial data, reduce coding requirements, and automate parts of GIS workflows. Related work has also explored structured prompting and schema alignment for planning and GIS tasks \citep{ying2026beyond}, code generation for transit data interaction \citep{devunuri2025transitgpt}, and the extraction of geospatial knowledge from language models for geographic prediction tasks \citep{manvi2023geollm}. Collectively, these efforts demonstrate the growing potential for LLMs to make GIS and transportation data systems more accessible to a wider range of users.

Many of these systems rely on direct code generation or agentic execution, which can offer flexibility but also introduces challenges related to non-determinism, lack of reproducibility, and error propagation into downstream outputs \citep{zhang2025geoanalystbench, qiu2025blueprint}. In more specialized analytical domains, these concerns have encouraged architectural approaches that separate natural language interpretation from downstream execution, relying instead on structured pipelines that operate independently of the language model \citep{jhamtani2023natural, barbieri2024llm, qiu2025blueprint}. These priorities align with broader expectations for trustworthy AI in public-sector settings. Frameworks such as the NIST AI Risk Management Framework \citep{nist_ai_rmf_2023} and its Generative AI Profile \citep{nist_genai_profile_2024} identify reliability, auditability, and human oversight as central requirements for consequential analytical systems. For NL interfaces to structured data systems, this means that design choices around schema conformance, validation, and interpretable execution are governance decisions as much as technical ones, since outputs need to be not only correct but traceable, verifiable, and consistent with the definitions, standards, and data practices that institutions and users rely on.

\subsection{Research Gap and Contribution}
\label{subsec:gap}
Existing work has made important progress in expanding NL access to transportation and geospatial data systems, and in establishing trustworthiness as a design requirement for public-sector AI. Transportation safety, however, remains a specialized planning and policy domain whose analytical requirements depend on domain-specific entities, field structures, and execution logic that general-purpose systems are not typically designed to address. Tasks such as proximity-based crash screening near schools or transit stops, infrastructure-linked prioritization, and program-specific temporal analysis call for structured, schema-grounded frameworks rather than open-ended query generation. At the same time, many of the stakeholders who rely on this type of analysis, including local agencies, school committees, and community advocates, may have limited expertise to navigate the technical workflows involved. To our knowledge, existing systems have not directly combined domain-specific transportation safety framing with NL accessibility in a way that supports reliable, reproducible analysis for broader non-specialist use. This gap is sociotechnical rather than purely computational: the key question is not only whether a language model can produce a spatial query, but whether GenAI-mediated access can be organized in a way that remains compatible with public-sector review, accountability, and planning practice.

This paper contributes a framework that uses generative AI as a controlled interface to structured transportation safety analysis, making it accessible to community members, advocates, municipal staff, and planning agencies who have safety questions but limited technical capacity for conventional GIS workflows. The framework is intentionally bounded to a domain-specific analytical schema, with language interpretation separated from execution through a transparent, rule-based validation layer that enforces schema conformance and produces auditable, reproducible outputs aligned with institutional planning requirements. This design serves both non-specialist users seeking accessible safety evidence and institutional users who need outputs that are inspectable and grounded in authoritative data. The framework is implemented on a statewide Massachusetts transportation safety database and evaluated on a structured set of queries covering the full range of supported analytical operations, with results discussed in terms of both system performance and practical implications for transportation safety planning and governance.

\section{System Architecture}
\label{sec:architecture}

\begin{figure}[htbp]
\centering
\includegraphics[width=1\textwidth]{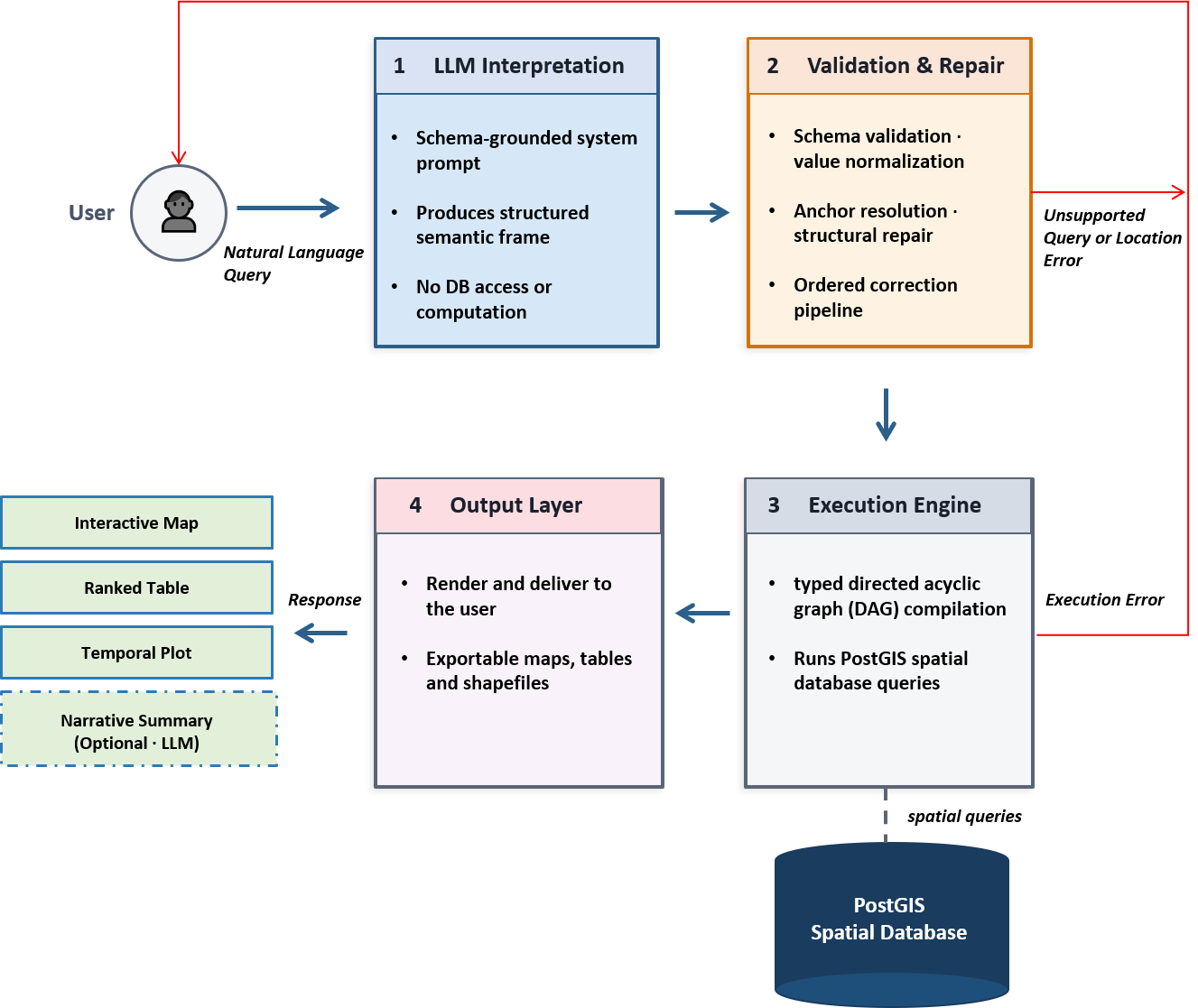}
\caption{System workflow from NL query to validated semantic frame, structured execution, and interactive output.}
\label{fig:architecture_overview}
\end{figure}

The system translates NL queries into structured spatial analyses through a multi-stage pipeline that separates language interpretation from analytical execution. User queries are first interpreted by an LLM into a semantic frame representing analytical intent. This frame is then processed by a Validation and Repair Layer that enforces schema conformance, normalizes values, and resolves geographic anchors before being compiled into a typed DAG of spatial operations and executed against the spatial database. Final outputs are presented through maps, tables, and related visualizations. Figure~\ref{fig:architecture_overview} summarizes the overall workflow.

\subsection{Study Area and Data}
\label{subsec:data}

The system is implemented on a statewide Massachusetts transportation safety database built on PostGIS, which integrates crash records from the Massachusetts Department of Transportation with roadway attributes and geospatial infrastructure layers. Crash records include attribute fields covering severity, first harmful event type, time of day, date, junction type, and sidewalk status, merged directly with roadway-level attributes to support infrastructure-linked filtering without additional joins. The database covers the full Massachusetts road network and all municipalities, providing statewide spatial coverage across urban, suburban, and rural environments.

The system operates on six entity types drawn from this database, summarized in Table~\ref{tab:entity_schema}. These entity types define the analytical schema: they determine what can be queried, how spatial relationships are constructed, and what attribute filters and ranking operations are supported. Crash severity is encoded across four canonical categories and the first harmful event field covers 30 categories drawn directly from the Massachusetts crash reporting standard, including collisions with pedestrians, cyclists, motor vehicles, fixed objects, and animals.

\begin{table*}[!htbp]
\centering
\caption{Supported entity types and key fields.}
\label{tab:entity_schema}
\scriptsize
\setlength{\tabcolsep}{3pt}
\renewcommand{\arraystretch}{1.1}
\begin{tabularx}{\textwidth}{@{}>{\raggedright\arraybackslash}p{0.12\textwidth}>{\raggedright\arraybackslash}p{0.10\textwidth}>{\raggedright\arraybackslash}p{0.12\textwidth}>{\raggedright\arraybackslash}X@{}}
\toprule
\textbf{Entity} & \textbf{Geometry} & \textbf{Records} & \textbf{Key fields} \\
\midrule
Crash     & Point   & 127,414$^*$ & Severity, first harmful event, crash date, crash time, sidewalk status (left/right), speed limit, junction type \\
\addlinespace
Road      & Line    & 504,905 & Speed limit, opposing direction speed limit, sidewalk status (left/right) \\
\addlinespace
School    & Point   & 2,448 & Name \\
\addlinespace
BusStop   & Point   & 17,933 & Stop ID, stop name \\
\addlinespace
Crosswalk & Polygon & 102,971 & ID \\
\addlinespace
Town      & Polygon & 333 & Name (used for geographic scoping) \\
\bottomrule
\end{tabularx}
\vspace{2pt}
\raggedright
\footnotesize
$^*$ Crash records cover the full calendar year 2025.
\end{table*}

\subsection{LLM Interpretation and Semantic Framing}
\label{subsec:llm_layer}

The LLM is used exclusively to interpret user queries. Each query is processed through a structured system prompt that defines supported entity types (Crash, Road, School, BusStop, Crosswalk, Town), their fields, valid values, supported spatial relationships, attribute operators, and role assignments.

The model produces a structured JSON representation that we refer to as a semantic frame. The term is used in the sense of task-oriented spoken language understanding, where an utterance is mapped to an intent plus a set of typed slots filled by entities and constraints \citep{tur2011spoken}, and more broadly in the tradition of frame semantics, where structured representations capture the participants and relations evoked by a scene \citep{baker1998framenet}. In our setting, the frame encodes analytical intent: which entities play which roles in the query (primary, support, scope, anchor, filter), what spatial and attribute constraints relate them, and how results should be ranked. Unlike linguistic semantic roles, which are tied to predicates, the roles here are analytical and tied to the operations supported by the execution engine. The frame thus serves as an intermediate representation between natural language and the typed DAG. At this stage, the semantic frame captures the model's initial interpretation but may still contain non-canonical expressions or structural inconsistencies that the validation layer will resolve. Figure~\ref{fig:validation_repair_example} (left) shows a representative example of a raw semantic frame as produced at this stage. The current implementation supports Gemini 2.5 Flash and GPT-4o as configurable options for the interpretation layer.

\subsection{Validation and Repair Layer}
\label{subsec:validation}

The Validation and Repair Layer serves as the intermediate governance layer between language interpretation and analytical execution. Its role is to transform the model’s approximate semantic frame into a schema-conformant representation suitable for structured analysis.

This layer performs four primary functions: schema validation, value normalization, anchor resolution, and structural correction. Schema validation checks entities, fields, and role assignments against the supported system registry. Value normalization converts natural language expressions into canonical database values. For example, \textit{``cyclists''} is normalized to \textit{``Collision with cyclist''}, \textit{``injury''} to \textit{``Non-fatal injury''}, and distance expressions such as \textit{``1km''} into internal numeric forms. Geographic references are resolved through geocoding or database lookup, while structural repair addresses incomplete or inconsistent analytical relationships.  Because this layer operates through rule-based correction logic, the boundary between language interpretation and structured execution remains stable regardless of how the upstream model expresses a given query.

\begin{figure}[htbp]
\centering
\includegraphics[width=1\textwidth]{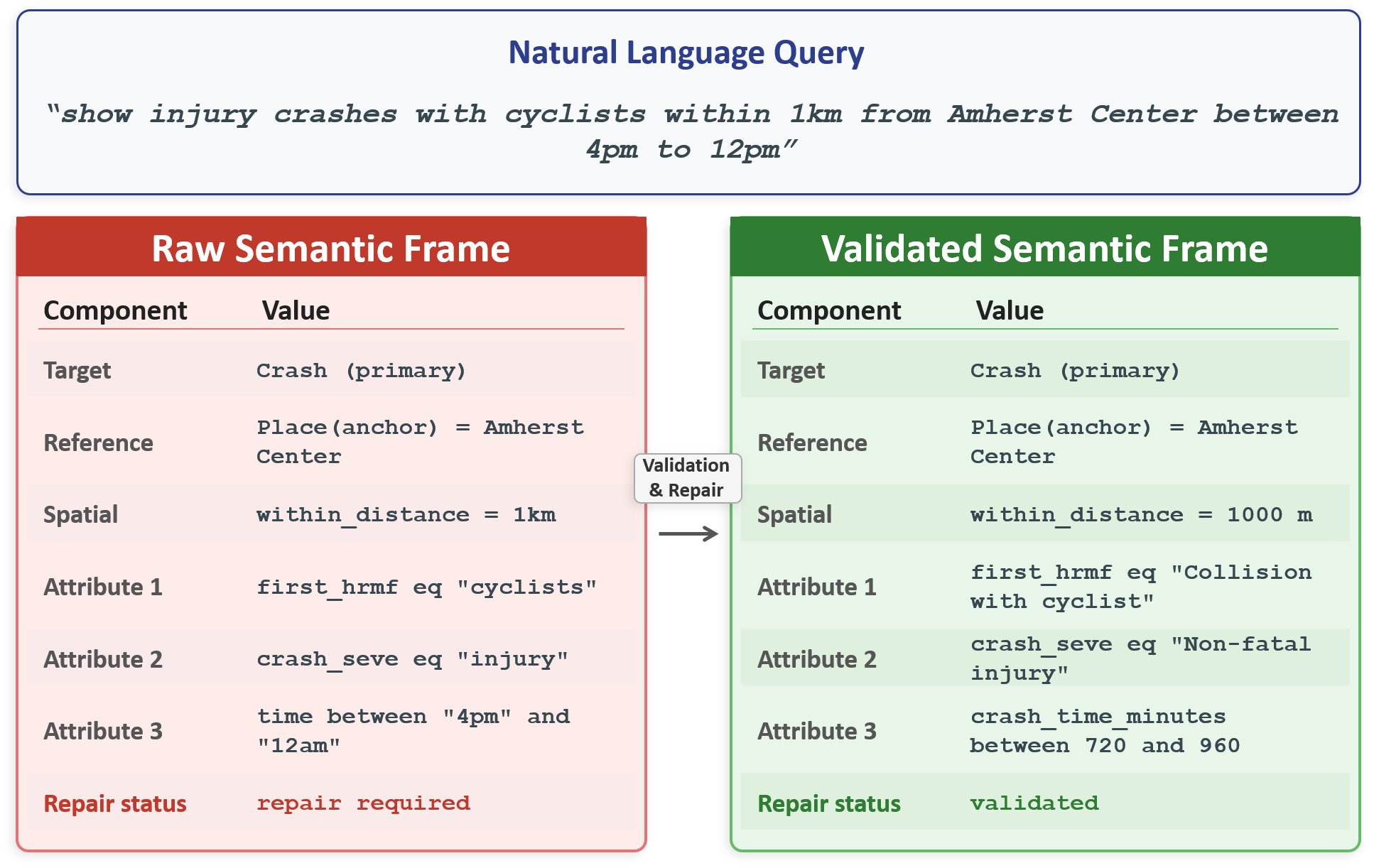}
\caption{Validation and repair example showing normalization of NL values into execution-ready representations.}
\label{fig:validation_repair_example}
\end{figure}

Figure~\ref{fig:validation_repair_example} illustrates this process for a representative query, showing how raw NL values are transformed into validated analytical specifications.

\subsection{Execution Engine and Output}
\label{subsec:execution}
Once validated and repaired, the semantic frame is compiled into a typed DAG of analytical operations and evaluated against the PostGIS spatial database. This design makes data dependencies between operations explicit and provides a reproducible and auditable pathway from validated intent to analytical output.

Each node in the execution graph represents a typed operation such as entity loading, attribute filtering, scope constraint application, spatial set matching, aggregation, or ranking. Edges between nodes encode data dependencies: a node executes only after all nodes it depends on have completed. The compiler determines this dependency structure directly from the validated semantic frame, so the graph topology reflects the analytical structure of the query.

\begin{figure}[htbp]
\centering
\includegraphics[width=0.9\textwidth]{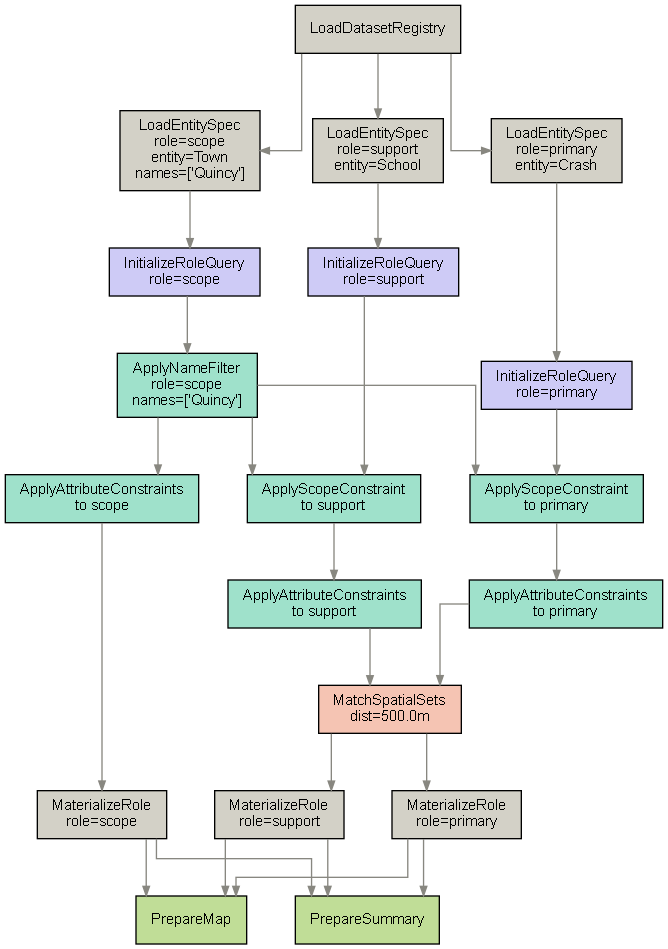}
\caption{Compiled execution DAG for the query \textit{``show crashes within 500m of all schools in Quincy.''}}
\label{fig:dag_example}
\end{figure}

The graph is also trackable during compilation: each input reference must correspond to an existing node, the graph must be acyclic, and the final nodes must correspond to valid output operations. These checks help identify structural problems in the compiled workflow before any query is sent to the database. Figure~\ref{fig:dag_example} illustrates the compiled execution graph for the query \textit{``show crashes within 500m of all schools in Quincy''}. Three independent entity-loading branches, Crash as primary, School as support, and Town as scope, fan out from a shared dataset registry, converge at scope constraint and spatial match nodes, and flow through role materialization to map and summary outputs.

\section{Evaluation}
\label{sec:evaluation}
\subsection{Evaluation Design}
\label{subsec:benchmark_design}
The system was evaluated using 80 NL queries organized into nine groups, each representing a different combination of supported analytical capabilities. G1 includes basic entity retrieval; G2 adds spatial scoping through towns, named places, or distance buffers; G3 introduces attribute filters such as crash severity, road user type, or roadway characteristics; G4 adds temporal constraints; and G5 tests spatial relationships between entity types. G6--G8 cover ranking tasks at the infrastructure, municipality, and road segment levels, while G9 combines multiple constraints within a single query. Query group definitions and representative prompts are provided in~\ref{app:benchmark_prompts}.

For each query, a ground truth entry was manually defined to specify the analytical constraints expected in the validated semantic frame, including entity roles, spatial relationships, attribute filters, temporal constraints, and ranking parameters where applicable. The ground truth represents the frame that should result after interpretation, validation, and repair, given the system's supported schema and operations. Evaluation was conducted at two levels. First, intent completeness assessed whether the validated semantic frame matched the expected ground truth. Second, execution success was assessed by whether the validated frame could be compiled into an execution graph and run against the database without error. The repair flag indicates whether the Validation and Repair Layer modified the raw LLM output before execution.

\subsection{Results}
\label{subsec:results}
The evaluation is interpreted within the system's defined operational scope: a bounded set of supported entity types, spatial relationships, attribute fields, and analytical operations grounded in the Massachusetts transportation safety database. Within this scope, all 80 queries executed successfully and all validated semantic frames matched their ground truth entries after validation and repair. Table~\ref{tab:group_results} summarizes execution times and repair counts by query group.

\begin{table}[htbp]
\centering
\small
\caption{Evaluation results and execution times by query group.}
\label{tab:group_results}
\begin{tabular}{@{}llrrrr@{}}
\toprule
Group & Category & $n$ & Avg (s) & Max (s) & Repaired \\
\midrule
G1 & Entity Retrieval     & 6  & 2.6   & 3.9    & 0 \\
G2 & Spatial Scoping      & 8  & 14.7  & 22.1   & 0 \\
G3 & Attribute Filter     & 12 & 11.4  & 21.8   & 3 \\
G4 & Temporal Filter      & 7  & 12.1  & 19.8   & 1 \\
G5 & Spatial Rel.         & 5  & 12.7  & 21.5   & 3 \\
G6 & Infra.\ Ranking      & 10 & 10.8  & 52.6   & 2 \\
G7 & Town Ranking         & 8  & 61.7  & 178.8  & 2 \\
G8 & Road Seg.\ Ranking   & 8  & 40.0  & 142.2  & 3 \\
G9 & Combined             & 16 & 9.3   & 20.8   & 9 \\
\midrule
   & \textbf{Overall} & \textbf{80} & \textbf{18.6} & \textbf{178.8} & \textbf{23} \\
\bottomrule
\end{tabular}
\end{table}

Overall, 23 of 80 queries (29\%) required correction by the validation layer before execution. Most corrections involved value normalization, mapping expressions such as \textit{``pedestrian''} to \textit{``Collision with pedestrian''} or \textit{``cyclist''} to \textit{``Collision with cyclist''}, accounting for 22 of the 25 individual repairs. The remaining three were structural, including the removal of a spurious anchor reference and the consolidation of a duplicate attribute constraint. The 29\% repair rate reflects the gap between NL expression and the requirements of schema-grounded execution.

Runtime was driven mainly by query structure and spatial scope. LLM interpretation took approximately 2--3 seconds per query regardless of complexity, while the remaining time came from database computation. Simple retrieval and filtering queries completed within 2--22 seconds. Ranking queries were more demanding: infrastructure ranking reached 53 seconds, town-level aggregation averaged 62 seconds, and road segment ranking reached 142 seconds for queries involving crash joins over municipal road networks. These runtimes reflect the cost of the underlying spatial operations, which are consistent with what equivalent analyses require in any GIS environment.

\section{Discussion}
\label{sec:discussion}

\subsection{Applications Across User and Decision-Making Contexts}
\label{subsec:applications}

Transportation safety concerns emerge in many different settings. A local school committee may want to better understand pedestrian crash patterns near one school, while a metropolitan planning organization may be focused on comparing municipalities for programs such as HSIP. These needs vary in scale, but both require accessible and structured safety analysis. This section demonstrates how the framework can support these different forms of work through two complementary analytical levels: localized safety diagnosis for site-specific concerns, and broader comparative screening for planning and prioritization. The examples are based on the Massachusetts implementation and reflect the current datasets, entities, and analytical operations supported by the system.

\subsubsection{Localized Safety Diagnosis and Community Evidence Generation}
\label{subsubsec:user_level}

At the site level, users are often trying to understand crash conditions and infrastructure characteristics around a specific school, bus stop, intersection, or any other location that matters in their community. These questions are usually driven by practical concerns. A parent group may be documenting pedestrian risks near a school, a municipality might be gathering evidence for a funding application, or a neighborhood organization may want to better understand the roadway environment around a transit stop. In these situations, proximity-based crash analysis and infrastructure visualization can be especially valuable, helping connect crash patterns and surrounding infrastructure conditions to the places people are most concerned about.

Local stakeholders often face practical challenges when trying to conduct these analyses themselves. Smaller agencies and municipalities may simply have more limited staff time and technical capacity for the GIS and data analysis work involved, which can make it harder to navigate complex safety datasets or federal funding processes\citep{fhwa_local_rural_funding}. At the same time, programs such as Safe Routes to School (SRTS) have noted that communities with the greatest safety needs, including those experiencing higher pedestrian crash burdens, are often also the least equipped to document those needs and compete for investment \citep{mcdonald2013assessing}. Reducing the technical barriers to site-specific safety analysis is therefore not just a matter of convenience. It can also influence whether local stakeholders are able to meaningfully participate in safety planning, advocacy, and funding opportunities.

\begin{figure}[htbp]
\centering
\includegraphics[width=0.9\textwidth]{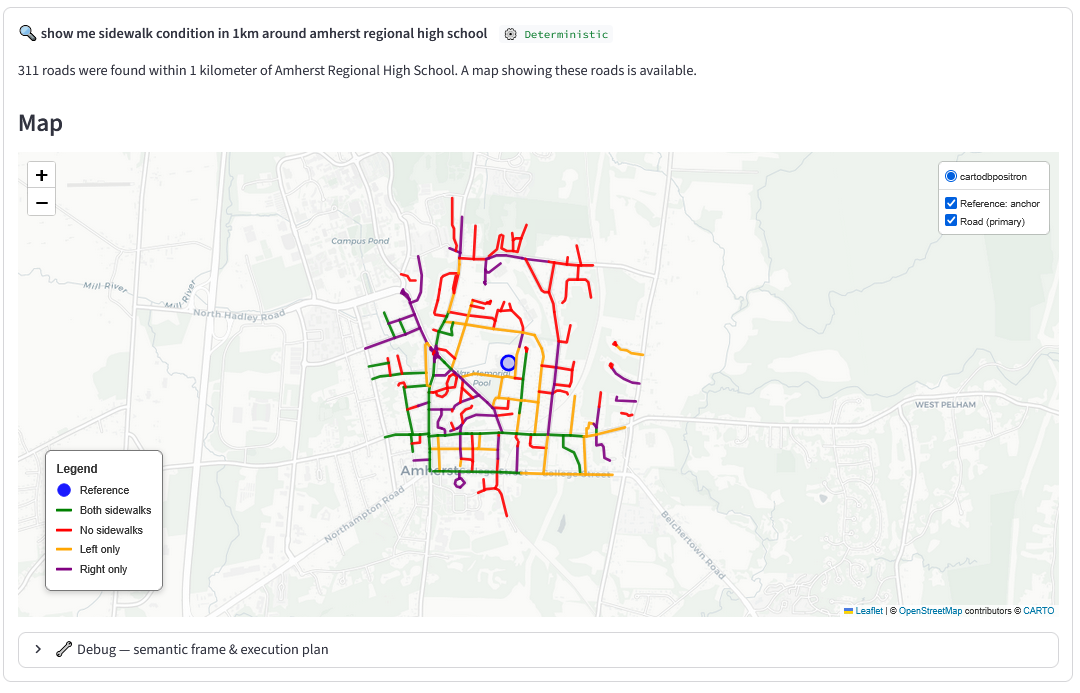}
\caption{System output for the query \textit{``show sidewalk conditions within 1km of Amherst Regional High School''}, showing mapped infrastructure attributes around a specific school site derived directly from the spatial database.}
\label{fig:sidewalk_amherst}
\end{figure}

Representative queries at this level include requests such as \textit{``show pedestrian crashes around Amherst Regional High School within 500m''}, \textit{``show crashes around Amherst Center within 1km''}, or \textit{``show crashes near Palmer St @ Brockton Ave bus stop''}. These types of questions generate location-centered maps and filtered crash records directly from the spatial database, allowing users to examine safety conditions around places of immediate local concern, including school zones, transit stops, and intersections where vulnerable road users face concentrated risk \citep{oke2025bus}. Infrastructure can also be evaluated in the same way through queries such as \textit{``show roads without sidewalks near bus stops in Amherst''} or as illustrated in Figure~\ref{fig:sidewalk_amherst}, which shows output for the query \textit{``show sidewalk conditions within 1km of Amherst Regional High School''}, where nearby sidewalk conditions are mapped around a specific school site. The system also supports targeted road segment screening, such as identifying high-risk corridors near individual schools or intersections.

\subsubsection{Decision-Maker-Level Comparative and Strategic Screening}
\label{subsubsec:decision_level}

At broader planning scales, the focus shifts from understanding conditions at a single location to comparing and prioritizing across many facilities, corridors, or jurisdictions. State DOTs, metropolitan planning organizations, transit agencies, and similar organizations often need to identify where crash exposure is highest, where infrastructure deficiencies are concentrated, or which locations may warrant funding or intervention priority \citep{fhwa_hsip_manual, fhwa_local_rural_funding}. At this level, the emphasis is on larger-scale screening, comparative evaluation, and resource allocation across broader transportation systems.

\begin{figure}[htbp]
\centering
\includegraphics[width=0.9\textwidth]{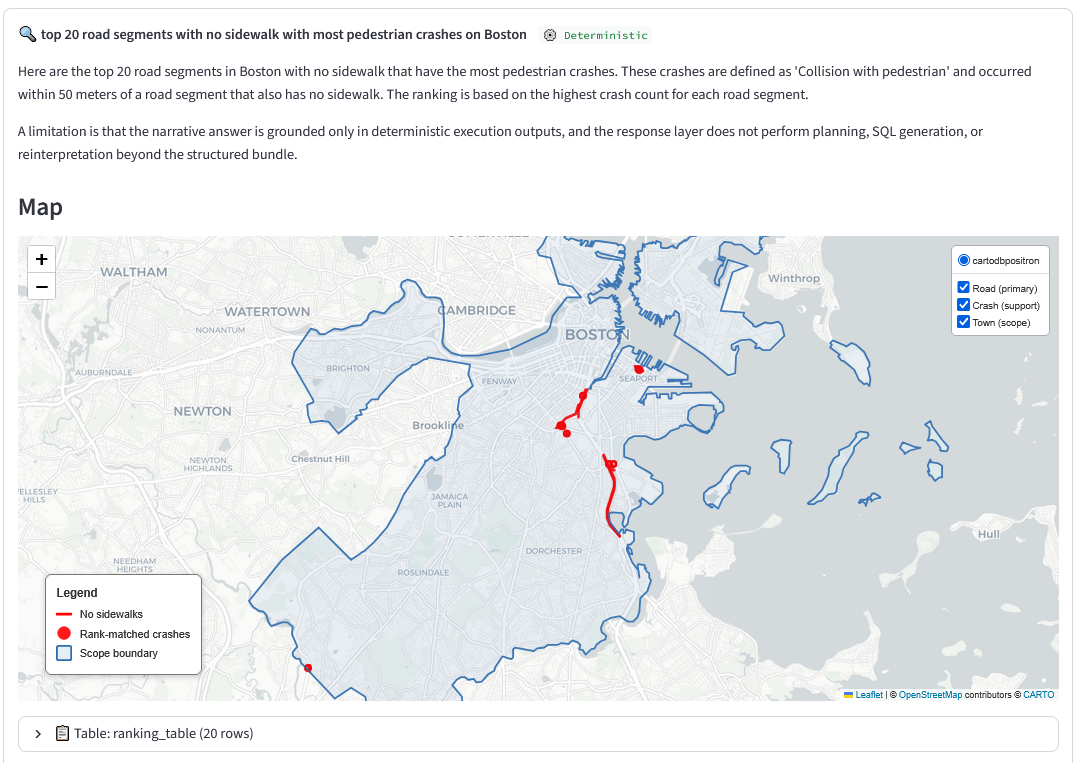}
\caption{System output for the query \textit{``top 20 road segments with no sidewalks on both
sides and the most pedestrian crashes''}, combining an infrastructure deficiency filter with
crash exposure ranking to identify corridors where pedestrian risk and missing infrastructure
coincide.}
\label{fig:roadseg_ranking}
\end{figure}

The framework supports these broader analyses through comparative ranking queries that combine crash records with spatial, infrastructure, and roadway filters across schools, bus stops, municipalities, and road segments. Queries such as \textit{``top 20 towns by pedestrian crashes''}, \textit{``top 10 schools by crashes within 500m in Boston''}, or \textit{``top 20 towns by crashes within 500m of bus stops''} allow users to compare safety conditions across larger geographic areas. These rankings can also incorporate more specific policy-relevant filters, including time of day, roadway speed limits, or facility type. For example, queries like \textit{``top 10 schools by crashes within 500m between 7am and 10am''} or \textit{``top 10 towns by crashes near bus stops with speed limit above 30''} can help decision-makers focus on particular risk dimensions that may align more closely with programmatic goals such as SRTS or HSIP \citep{mcdonald2013assessing, fhwa_hsip_manual}. At the corridor level, the same framework can evaluate crash burdens alongside infrastructure deficiencies. Figure~\ref{fig:roadseg_ranking} provides one example through the query \textit{``top 20 road segments with no sidewalks on both sides and the most pedestrian crashes''}, identifying roads where missing pedestrian infrastructure and elevated crash exposure overlap.

In its current implementation, ranking metrics primarily reflect crash counts under user-specified spatial and attribute constraints. However, agencies that rely on weighted severity measures, established scoring systems, or other composite prioritization frameworks can adapt the ranking logic within the semantic structure to match their own institutional criteria. This flexibility allows the system to serve as a structured execution interface for organization-specific decision processes rather than imposing a single universal metric. At the same time, because the framework applies consistent schema definitions and deterministic execution logic, results remain reproducible across users and sessions. This consistency is particularly important in institutional settings where analyses may be reviewed by technical staff, compared across jurisdictions, or used to support formal prioritization and funding decisions.

\subsection{Trustworthiness Considerations}
\label{subsec:governance}
As NL interfaces for analytical systems become more visible in public-sector planning, questions of trustworthiness and governance become increasingly relevant. Comparative reviews of government GenAI guidance consistently identify hallucination, over-reliance, bias, and data privacy as central concerns in public-sector deployment \citep{beltran2024comparative, nist_genai_profile_2024}. Table~\ref{tab:risks} examines these risks in the context of this framework, summarizing how the design addresses each and what limitations remain.

\begin{table*}[!htbp]
\centering
\caption{Risks of GenAI in public-sector analytical systems, design mitigations, and remaining limitations.}
\label{tab:risks}
\scriptsize
\setlength{\tabcolsep}{3pt}
\renewcommand{\arraystretch}{1.2}
\begin{tabularx}{\textwidth}{@{}>{\raggedright\arraybackslash}p{0.20\textwidth}>{\raggedright\arraybackslash}X>{\raggedright\arraybackslash}X@{}}
\toprule
\textbf{Risk} & \textbf{Mitigation} & \textbf{Limitations and remaining risks} \\
\midrule
Hallucination and output validity & The language model handles interpretation only; analytical logic runs deterministically from a validated semantic frame, so reasoning errors cannot propagate into outputs. & The frame may still misrepresent intent if an ambiguous query is interpreted plausibly but incorrectly. An NL summary surfaced with results allows users to verify whether their question was correctly understood. \\
\addlinespace
Non-reproducibility across users and sessions & The same validated frame always produces the same output against the same database state, independent of model temperature or session context. & Reproducibility holds within the current schema. Database changes or validation rule updates may affect results over time if not versioned. \\
\addlinespace
Lack of auditability & The semantic frame and execution graph provide an inspectable record of all analytical steps, reviewable independently of the query. & Meaningful review of the semantic frame and execution graph still requires some technical familiarity. \\
\addlinespace
Over-reliance on outputs & Outputs are framed as decision-support evidence for human review. The system generates no interpretive conclusions or policy recommendations. & The design cannot prevent over-reliance. Users may treat outputs as authoritative without verifying query logic, particularly under time pressure. \\
\addlinespace
Unequal access reproducing disparities & Removing GIS and database prerequisites lowers barriers for smaller agencies, community organizations, and non-specialist stakeholders. & Lowering technical barriers does not address data gaps. Communities with greater safety needs may be underrepresented in crash or infrastructure data, limiting what outputs can reflect. \\
\addlinespace
Privacy and data governance & Queries are constrained to schema-defined fields; the language model never accesses individual records directly, and execution stays within the controlled pipeline. & It remains the responsibility of jurisdictions to keep private data out publicly accessible datasets or data fields. \\
\addlinespace
Fairness and equity in prioritization & Consistent spatial definitions and execution logic are applied uniformly across all queries and jurisdictions. & Consistent execution does not guarantee equitable representation in outcomes. For example, ranking towns or locations by risk depends on the defined schema, which does not necessarily include all relevant data for fair decision making. \\
\bottomrule
\end{tabularx}
\end{table*}

The table suggests that the bounded design addresses several of the most critical risks associated with GenAI in public-sector analytical settings, particularly around reproducibility, auditability, and hallucination propagation. The remaining limitations in each case point to conditions that system design alone cannot resolve, including how analytical concepts such as risk and priority are defined during deployment, how outputs are interpreted under institutional pressure, and how underlying data quality shapes what the framework can surface. Trustworthy deployment therefore depends not only on architectural choices but also on the institutional practices and governance arrangements that surround them.

\subsection{Future Development Pathways}
\label{subsec:future_work}

The current implementation is intentionally bounded: it supports a defined set of entity types, spatial relationships, attribute fields, and analytical operations grounded in the Massachusetts transportation safety database. This scope reflects a deliberate design choice to maintain schema-grounded execution within a well-understood domain, but it also points toward several natural directions for extension.

The most straightforward path is expanding the supported analytical vocabulary. Because interpretation and execution are separated, new entity types, attribute fields, and analytical operations can be added modularly without restructuring the pipeline. This could include rate-based and exposure-adjusted screening metrics, severity-weighted ranking, network-based accessibility measures, and more flexible temporal aggregation. Extending the schema to incorporate additional data layers such as pedestrian volume counts, land use, or demographic vulnerability indicators would also allow for more equity-sensitive analyses, addressing one of the limitations noted in the governance discussion.

Another related direction is adapting the framework to specific institutional workflows. The current system functions as a general transportation safety interface, but the same architecture could be configured around particular planning contexts such as SRTS screening, HSIP network analysis, transit access evaluation, or corridor prioritization. Tailoring the system prompt, supported operations, and output formats to these specific workflows could improve practical usability for the agencies and organizations most likely to deploy such tools.

Scaling the framework to other jurisdictions and database schemas is a longer-term challenge. The system's reliability depends on the quality and consistency of schema grounding, which in turn depends on the underlying data model. Deploying the framework in new jurisdictions would require schema adaptation and likely a new round of validation tuning. How much of this process can be automated, and how much depends on domain expertise, is an open question.

Finally, the current system handles queries as independent one-shot requests. Many real planning workflows are iterative, involving follow-up questions, constraint refinement, and comparison across alternatives. Supporting multi-turn interaction, clarification of ambiguous references, and structured comparison across query variants would bring the interface closer to how analysts actually work. Alongside technical development, user-centered evaluation with planners, municipal staff, and community practitioners would help assess not only system performance but also practical accessibility and decision-support value in real institutional settings.
\section{Conclusion}
\label{sec:conclusion}

Transportation safety analysis has become increasingly data-driven, but the technical workflows it depends on remain difficult to access for many of the stakeholders who need them most. Local agencies, school committees, community organizations, and planners often have clear safety questions but limited capacity to navigate the GIS platforms, database systems, and spatial analysis tools required to answer them. This paper describes a framework with an NL interface designed to narrow that gap by allowing users to query an authoritative transportation safety database in plain language and receive structured analytical outputs directly.

The system works by separating language interpretation from analytical execution. A language model translates user queries into structured semantic frames, which are then validated and corrected by a rule-based layer before being compiled into deterministic spatial operations against a statewide Massachusetts crash and infrastructure database. The 29\% repair rate observed in the evaluation is the most concrete finding: nearly a third of queries required correction before execution could proceed, reflecting the real gap between flexible NL and the strict requirements of schema-grounded analysis. An important contribution of this paper is the proposed design principle. Separating interpretation from execution, and placing a rule-based validation layer at the boundary between them, is a practical approach to integrating generative AI into analytical workflows where reproducibility, auditability, and institutional trust matter. Whether that principle scales to other safety domains, other jurisdictions, or more complex analytical tasks remains to be seen, but the current results suggest it is a viable direction worth pursuing.

\section{Declaration of generative AI and AI-assisted technologies in the manuscript preparation process}

During the preparation of this work the authors used generative AI tools in the following ways: 1) Gemini 2.5 Flash and GPT-4o as configurable options for the interpretation layer as described in Section~\ref{subsec:llm_layer}; 2) Claude Opus was used to support the development of the case study code; and 3) Claude Sonnet was used to edit grammar and readability in the manuscript. After using these, the authors reviewed and edited the content as needed and take full responsibility for the content of the published article.

\appendix
\section{Semantic Frame Schema and Example}
\label{app:semantic_frame}
Each NL query is interpreted by the language model into a structured
JSON object called a semantic frame. The frame encodes the full analytical intent
of the query before any execution takes place. Table~\ref{tab:frame_schema}
summarizes the six top-level components of the frame schema.

The following example shows the validated semantic frame produced for the query
\textit{``top 5 schools by pedestrian crashes within 500m in Boston''}:

\begin{verbatim}
{
  "supported": true,
  "targets": [
    {"entity": "School", "role": "primary"},
    {"entity": "Crash",  "role": "support"},
    {"entity": "Town",   "role": "scope"}
  ],
  "references": [
    {"entity": "Town", "role": "scope", "name": "Boston"}
  ],
  "spatial_constraints": [
    {"relation": "within_distance",
     "target_role": "support",
     "reference_role": "primary",
     "distance_m": 500.0}
  ],
  "attribute_constraints": [
    {"target_role": "support",
     "field": "first_hrmf",
     "operator": "eq",
     "value": "Collision with pedestrian"}
  ],
  "relations": [],
  "ranking": {
    "metric": "crash_count",
    "target_role": "primary",
    "order": "highest",
    "top_n": 5
  }
}
\end{verbatim}

\begin{table*}[!htbp]
\centering
\caption{Semantic frame schema components.}
\label{tab:frame_schema}
\scriptsize
\setlength{\tabcolsep}{3pt}
\renewcommand{\arraystretch}{1.3}
\begin{tabularx}{\textwidth}{@{}>{\raggedright\arraybackslash}p{0.22\textwidth}>{\raggedright\arraybackslash}X@{}}
\toprule
\textbf{Component} & \textbf{Description} \\
\midrule
\texttt{targets} & Entity role assignments. Each target specifies an entity type and one of five roles: \textit{primary} (displayed or ranked), \textit{support} (aggregation measure), \textit{scope} (geographic boundary), \textit{anchor} (geocoded reference point), or \textit{filter} (spatial pre-filter). \\
\addlinespace
\texttt{references} & Named geographic references such as school names or place names, resolved to verified spatial coordinates during validation. \\
\addlinespace
\texttt{spatial\_constraints} & Geometric relationships between roles, specifying a relation type (\texttt{within\_distance}, \texttt{intersects}, \texttt{contains}, \texttt{nearest\_to}), connected roles, and an optional distance parameter. \\
\addlinespace
\texttt{attribute\_constraints} & Thematic filters on entity fields, each specifying a target role, field name, operator, and value. Supported operators include \texttt{eq}, \texttt{in}, \texttt{gt}, \texttt{gte}, \texttt{lt}, \texttt{lte}, \texttt{between}, \texttt{is\_null}, and \texttt{not\_null}. \\
\addlinespace
\texttt{relations} & Semantic links between entity pairs, such as snap-matching crashes to road segments. \\
\addlinespace
\texttt{ranking} & When applicable, specifies the aggregation metric, the role to rank, ordering direction, and result count. \\
\bottomrule
\end{tabularx}
\end{table*}

\section{Supported Entities and Schema}
\label{app:schema}
The system supports six entity types drawn from the statewide Massachusetts transportation safety database. Table~\ref{tab:entity_schema} summarizes each entity, its geometry type, and its key analytical fields.

Crash severity takes one of four canonical values: \textit{Property damage only}, \textit{Non-fatal injury}, \textit{Fatal injury}, and \textit{Unknown}. The first harmful event field supports 30 canonical categories including collisions with pedestrians, cyclists, motor vehicles, fixed objects, and animals, drawn directly
from the Massachusetts crash reporting standard.

\section{Validation and Repair: Normalization Examples}
\label{app:repair}
The Validation and Repair Layer maps free-form NL expressions to canonical schema values before execution. Table~\ref{tab:normalization} shows representative examples of corrections applied during the evaluation.

\begin{table*}[!htbp]
\centering
\caption{Representative normalization corrections applied by the Validation and Repair Layer.}
\label{tab:normalization}
\scriptsize
\setlength{\tabcolsep}{3pt}
\renewcommand{\arraystretch}{1.1}
\begin{tabularx}{\textwidth}{@{}>{\raggedright\arraybackslash}p{0.18\textwidth}>{\raggedright\arraybackslash}X>{\raggedright\arraybackslash}X@{}}
\toprule
\textbf{Field} & \textbf{Raw LLM value} & \textbf{Normalized value} \\
\midrule
First harmful event & \texttt{cyclists} & \texttt{Collision with cyclist} \\
\addlinespace
First harmful event & \texttt{pedestrian} & \texttt{Collision with pedestrian} \\
\addlinespace
First harmful event & \texttt{bike} & \texttt{Collision with cyclist} \\
\addlinespace
Crash severity & \texttt{injury} & \texttt{Non-fatal injury} \\
\addlinespace
Crash severity & \texttt{fatal} & \texttt{Fatal injury} \\
\addlinespace
Crash severity & \texttt{pdo} & \texttt{Property damage only (none injured)} \\
\addlinespace
Distance & \texttt{1km} & \texttt{1000 m} \\
\addlinespace
Distance & \texttt{half a mile} & \texttt{804 m} \\
\addlinespace
Time & \texttt{between 4pm and 8pm} & \texttt{[960, 1200] minutes} \\
\addlinespace
Ranking order & \texttt{most} & \texttt{highest} \\
\addlinespace
Ranking order & \texttt{fewest} & \texttt{lowest} \\
\bottomrule
\end{tabularx}
\end{table*}

Geographic anchor resolution follows normalization. School names and place names are resolved to verified spatial coordinates through database lookup or geocoding before execution proceeds. Where multiple candidate locations are found for a place name, the system surfaces the options to the user for selection rather than proceeding with an ambiguous reference.

\clearpage
\section{Query Groups and Representative Prompts}
\label{app:benchmark_prompts}
Table~\ref{tab:benchmark_groups} summarizes the nine query groups and the
analytical capability combinations each represents. Table~\ref{tab:benchmark_prompt_examples}
provides representative prompts from each group.

\begin{table}[htbp]
\centering
\footnotesize
\caption{Query groups and the analytical capability combinations represented in the evaluation.}
\label{tab:benchmark_groups}
\setlength{\tabcolsep}{4pt}
\renewcommand{\arraystretch}{1.1}
\begin{tabular}{@{}lccccccc@{}}
\toprule
Group ($n$) & Ret. & Scope & Attr. & Temp. & Spatial & Rank & Comp. \\
\midrule
G1 (6)  & \checkmark &            &            &            &            &            &            \\
G2 (8)  & \checkmark & \checkmark &            &            & \checkmark &            &            \\
G3 (12) & \checkmark & \checkmark & \checkmark &            &            &            &            \\
G4 (7)  & \checkmark & \checkmark & \checkmark & \checkmark &            &            &            \\
G5 (5)  & \checkmark & \checkmark & \checkmark &            & \checkmark &            &            \\
G6 (10) & \checkmark & \checkmark & \checkmark & \checkmark & \checkmark & \checkmark & \checkmark \\
G7 (8)  & \checkmark &            & \checkmark & \checkmark & \checkmark & \checkmark & \checkmark \\
G8 (8)  & \checkmark & \checkmark & \checkmark &            & \checkmark & \checkmark & \checkmark \\
G9 (16) & \checkmark & \checkmark & \checkmark & \checkmark & \checkmark & \checkmark & \checkmark \\
\bottomrule
\end{tabular}

\vspace{4pt}
\begin{minipage}{\linewidth}
\footnotesize
\textit{Note:} Ret. = entity retrieval; Attr. = attribute filtering; Temp. = temporal filtering; Comp. = combined multi-constraint queries.
\end{minipage}
\end{table}

\begin{table*}[!htbp]
\centering
\caption{Representative evaluation prompts by group.}
\label{tab:benchmark_prompt_examples}
\scriptsize
\setlength{\tabcolsep}{3pt}
\renewcommand{\arraystretch}{1.1}
\begin{tabularx}{\textwidth}{@{}>{\raggedright\arraybackslash}p{0.08\textwidth}>{\raggedright\arraybackslash}X@{}}
\toprule
\textbf{Group} & \textbf{Example prompts} \\
\midrule
G1 & show crashes in Quincy; show roads in Amherst \\
\addlinespace
G2 & show crashes around Amherst Center within 1km; show schools within 1km around Amherst Center \\
\addlinespace
G3 & show fatal crashes in Quincy; show crashes with speed limit above 30 in Quincy \\
\addlinespace
G4 & show crashes between 7am and 10am in Quincy; show crashes between January 6 2025 and February 5 2025 \\
\addlinespace
G5 & show crashes near crosswalks in Amherst; show crashes near bus stops in Quincy \\
\addlinespace
G6 & top 10 schools by crashes within 500m in Quincy; top 10 schools by crashes within 500m between 7am and 10am \\
\addlinespace
G7 & top 20 towns by crashes; top 10 towns by crashes within 500m of schools \\
\addlinespace
G8 & top 10 road segments by pedestrian crashes in Amherst; top 10 road segments by crashes without sidewalks in Amherst \\
\addlinespace
G9 & top 10 schools by pedestrian crashes within 500m in Quincy between 7am and 10am; show fatal pedestrian crashes in Quincy \\
\bottomrule
\end{tabularx}
\end{table*}

\bibliographystyle{elsarticle-harv}
\bibliography{cas-refs}

\end{document}